\pdfoutput=1
\documentclass[11pt]{article}

\usepackage[preprint]{acl}

\usepackage{times}
\usepackage{latexsym}

\usepackage{tikz}
\usetikzlibrary{arrows.meta, positioning, shapes.multipart, fit, shapes.geometric}
\tikzset{
  block/.style = {rectangle, rounded corners, draw, align=left, fill=gray!5, minimum width=4.5cm},
  decision/.style = {diamond, draw, aspect=2, align=center, fill=gray!5, inner sep=1pt},
  startstop/.style = {ellipse, draw, align=center, fill=gray!10},
  line/.style = {->, >=Latex},
  smallblock/.style = {rectangle, draw, rounded corners, align=left, fill=gray!5, minimum width=4.5cm, font=\footnotesize},
  textnode/.style = {align=left}
}

\usepackage[T1]{fontenc}
\usepackage[utf8]{inputenc}
\usepackage{booktabs}
\usepackage{tabularx}
\usepackage{comment}
\usepackage{xcolor}               % For color
\usepackage{inconsolata}

\usepackage{graphicx}

\newcommand{\coloredcitep}[2]{\begingroup\hypersetup{citecolor=#1}\citep{#2}\endgroup}

\newcommand{\mysection}{\vspace*{-1.5mm}\section}
\newcommand{\mysubsection}{\vspace*{-1mm}\subsection}

\newcommand{\myparagraph}{\vspace*{-2mm}\paragraph}

\newcommand{\eg}{e.g., }

\title{From Instruction to Output: The Role of Prompting in Modern NLG}

  \author{
  \textbf{Munazza Zaib\textsuperscript{1,2}},
  \textbf{Elah Alhazmi\textsuperscript{3}}
\\
\\
  \textsuperscript{1}Faculty of Information Technology, Monash University, Melbourne, Australia\\
  \textsuperscript{2}School of Computer, Data and Mathematical Sciences, Western Sydney University, Sydney, Australia\\
  \textsuperscript{3}School of Computing, Macquarie University, Sydney, Australia\\
\\
  \small{
    \textbf{Correspondence:} \href{mailto:munazza.zaib@monash.edu}{munazza.zaib@monash.edu},
    \href{mailto:elaf.alhazmi@hdr.mq.edu.au}{elaf.alhazmi@hdr.mq.edu.au}
  }
}

\begin{document}
\maketitle
\begin{abstract}
Prompt engineering has emerged as an integral technique for 
extending the strengths and abilities of Large Language Models 
(LLMs) to gain significant performance gains in various Natural 
Language Processing (NLP) tasks. This approach, which requires 
instructions to be composed in natural language to bring out 
the knowledge from LLMs in a structured way, has driven 
breakthroughs in various NLP tasks. Yet, there is still no 
structured framework or coherent understanding of the varied 
prompt‐engineering methods and techniques, particularly in 
the field of Natural Language Generation (NLG). 

This survey aims to help fill that gap by outlining 
recent developments in prompt engineering, and their effect on different NLG tasks. It reviews recent advances in prompting methods and their impact on NLG tasks, presenting prompt design as an input-level control mechanism that complements fine-tuning and decoding approaches.
The paper introduces a taxonomy of prompting paradigms, a decision framework for prompt selection based on varying factors for the practitioners,  outlines emerging trends and challenges, and proposes a framework that links design, optimization, and evaluation to support more controllable and generalizable NLG.

\end{abstract}

\mysection{Introduction}
\vspace*{-1mm}
The rapid progress of artificial intelligence (AI) in recent years has been driven largely by large language models (LLMs), which achieve strong performance across many Natural Language Processing (NLP) tasks, including question answering, storytelling, summarization, machine translation, and sentiment analysis \cite{PromptEng2024ShubhamEtAl}.
%%, and dialogue generation. 
Within this broader NLP landscape, Natural Language Generation (NLG) 
%%remains a particularly challenging and essential subdomain as it requires models not only to understand language but also to produce coherent, contextually appropriate, and controllable outputs. 
poses its own challenges. Despite the fluency and versatility of LLMs, 
generating high-quality text for diverse NLG tasks often demands 
careful guidance that goes beyond improvements in model 
architecture~\cite{NLG2018GattEtAl,SequentialNLG2023PandeyEtAl} or 
task-specific fine-tuning~\cite{Parameter2023XuEtAl}. In this context, 
prompt engineering~\cite{Pretrain2023LiuEtAl} has emerged as a 
promising paradigm for steering LLM outputs in NLG, enabling flexible 
control over style, structure and content, without additional retraining. 
Prompt engineering is a technique whereby natural language instructions, or prompts, are used to guide an LLM's behavior responses, with the aim of improving accuracy,  relevance and coherence in the generated output~\cite{Unleashing2025ChenEtAl}. 
This approach offers a practical, low-resource alternative for advancing 
NLG applications, making it increasingly relevant as the field moves towards building robust and adaptable generation systems.

Prompt design plays a critical role in shaping the structure and coherence 
of a response across a wide range of NLG tasks~\cite{ThePromptReprt2024SchulhoffEtAl}. Owing to its use of 
instructions and examples written in natural language, it constitutes a
bridge between users and LLMs, letting users decide and guide an LLM's behavior. Despite its increasing popularity and significance, prompt engineering remains underrepresented in existing surveys on NLG. Most of the surveys focus on 
innovation in architecture designs~\cite{NLGSequentialArchitectures2023PandeyEtAl,NLGDecoding2021ZarriessEtAl},  evaluation methods~\cite{EvaluationMetric2005StentEtAl, MetricsEvaluationSurvey2022SaiEtAl, EvaluationTextGeneration2020CelikyilmazEtAl}, or the categorization of downstream  tasks~\cite{NLG2023DongEtAl, NLG2018GattEtAl, DialogGenerationSurvey2019SanthanamEtAl}. 
As a result, there is a need for a survey focusing just on prompting  for NLG and its applications. 

This survey addresses that gap by providing a structured synthesis of prompt engineering for NLG and its applications. Concretely, we (i) contrast prompt-based control with fine-tuning and decoding-level control, (ii) introduce a taxonomy of prompting paradigms for NLG, (iii) analyse how prompts influence key control dimensions such as content/factuality, structure/length, and style/tone, and (iv) review evaluation practices, robustness issues, and emerging trends, culminating in a systematic framework that links design, optimization, and evaluation.

\subsection{Positioning and Distinct Contributions}
\label{sec:positioning}

Recent surveys have substantially advanced the understanding of prompting, but they differ in scope and organizing principles from the objectives of this work. 
\citet{Pretrain2023LiuEtAl} provide a unifying view of \emph{prompt-based learning} across NLP, introducing formal notation and organizing prior work along dimensions such as model choice, prompt/template design, and tuning strategies. 
In contrast, \citet{ThePromptReprt2024SchulhoffEtAl} compile a large-scale taxonomy and vocabulary of prompt engineering techniques (including multimodal variants), and discuss broad issues such as benchmarking, security, and evaluation practices. 
Similarly, \citet{Unleashing2025ChenEtAl} review prompt engineering for LLMs and extend the discussion to vision language models, adversarial attacks, and evaluation.

\paragraph{How this survey differs?}
This survey focuses specifically on \textbf{prompting for Natural Language Generation (NLG)} and treats prompting as an \textbf{input-level control mechanism} for steering NLG outputs along multiple control dimensions, complementing fine-tuning and decoding-level interventions.
Rather than primarily cataloging techniques, we organize the literature around an actionable \textbf{control pipeline} that connects: 
(i) \emph{prompt design choices} (instruction structure, roles, exemplars, constraints), 
(ii) \emph{prompt optimization strategies} (manual refinement, search-based methods, and hybrid approaches), and 
(iii) \emph{evaluation and robustness} considerations that are uniquely pronounced in prompted NLG (sensitivity to phrasing, reasoning depth, and context handling).

\begin{figure}[tb!]
    \centering
    \includegraphics[width=\linewidth]{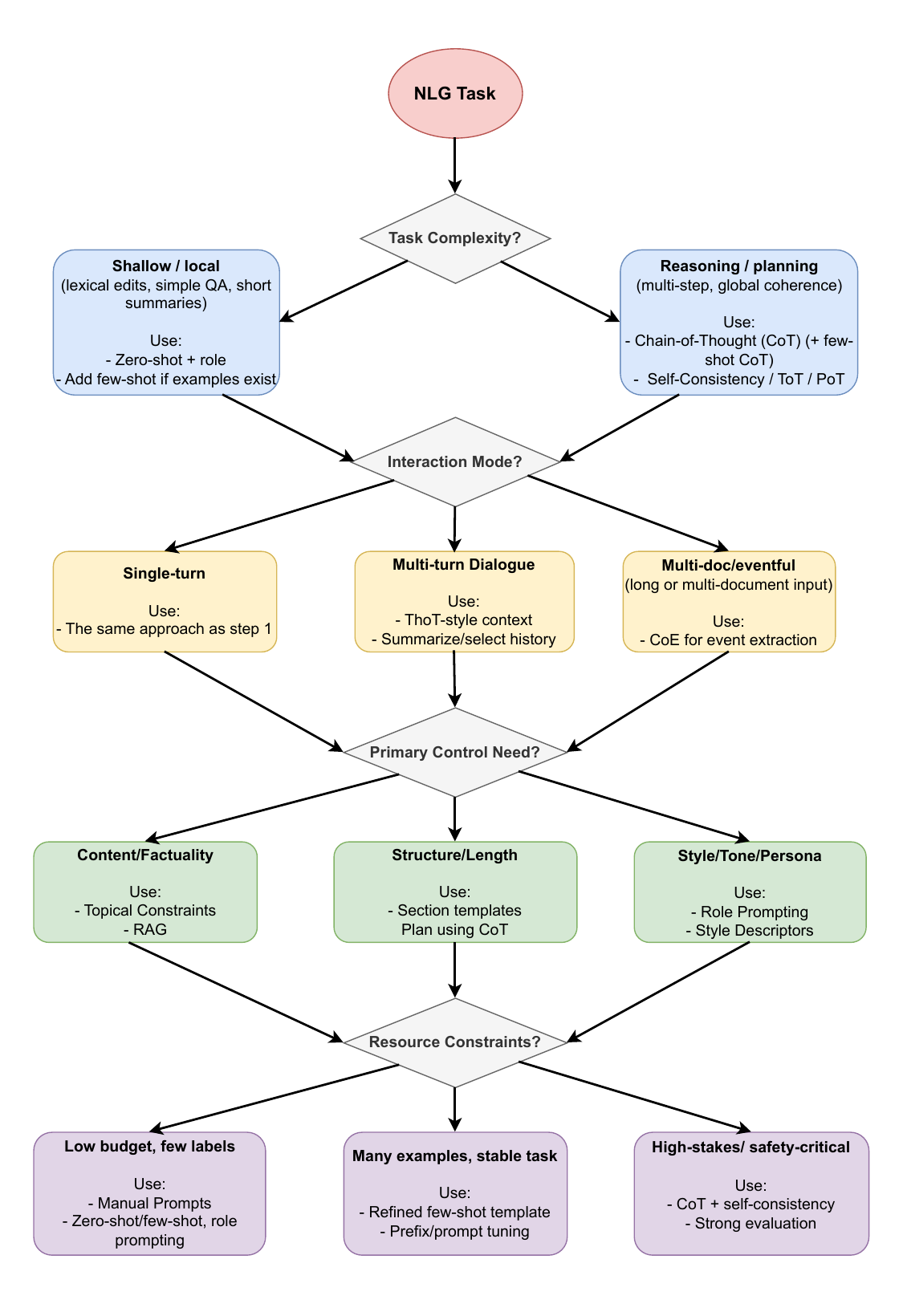}
    \caption{Decision framework for selecting prompt-engineering strategies for NLG based on task complexity, interaction setting, primary control objective, and resource constraints.}
    \label{fig:flowchart}
\end{figure}

\paragraph{What readers gain?}
This NLG-centric perspective provides two practitioner-focused contributions. First, we offer task-grounded guidance for mapping prompting paradigms to common NLG settings via a taxonomy that distinguishes \emph{foundational}, \emph{contextual}, and \emph{advanced reasoning} paradigms. Second, we operationalise this guidance as a decision framework (Figure~\ref{fig:flowchart}) that supports prompt strategy selection based on task complexity, interaction setting, primary control objective, and resource constraints. Finally, we consolidate these insights into a systematic framework that formalises prompt engineering through three interdependent dimensions, namely \emph{design}, \emph{optimization}, and \emph{evaluation}, to help make prompting less of a ``trial-and-error'' activity and more of a systematic, reliable way to control what the model generates

\mysection{Comparison with Fine-Tuning and Decoding-Level Control}
\vspace*{-1mm}
Prompt-based control operates at the input level, requiring
no extra training and providing fast adaptability to 
new control goals. In contrast, fine-tuning allows deeper 
integration of control signals by aligning an LLM’s internal 
representations with desired outputs. This achieves higher
consistency than prompt-based NLG outputs,
but at higher computational and data costs~\cite{PromptvsFineTuning2025ShinEtAl}. 

Decoding-level control (\eg constrained beam search, nucleus 
sampling adjustments) \cite{Decoding2023Naseh2023} manipulates the
generation process after an LLM’s next-token probabilities 
are computed, enabling some lexical or length constraints,
but with limited flexibility for high-level aspects of generation, such as discourse structure, tone, or content framing \cite{NeuralTextDegeneration2020HoltzmanEtAl}.

Thus, prompt engineering holds a unique middle position. It is more cost-effective and adaptable than fine-tuning, while 
offering broader control dimensions than decoding-level control.
This makes prompt engineering an effective strategy across various 
NLG tasks such as summarization, story generation, dialogue generation etc. \cite{PromptEng2024ShubhamEtAl}.

Findings in recent studies both substantiate the functional distinctions between prompting, fine-tuning, and decoding-level methods, and show a growing convergence through hybrid approaches that integrate elements of each. Parameter-efficient methods such as prefix tuning \cite{PrefixTuning2021Li} and prompt tuning \cite{PromptTuning2021Lester} demonstrate that lightweight fine-tuning can approach the performance of full model adaptation, offering a middle ground between static prompts and full retraining. 
Similarly, decoding-level techniques such as NeuroLogic Decoding \cite{Lu2021NeuroLogic}, contrastive search \cite{Su2022Contrastive}, and contrastive decoding \cite{ShiContrastiveDecoding2023} enhance coherence, factual grounding, and stylistic alignment beyond conventional constrained beam search. 
At the intersection of these paradigms, interdisciplinary approaches such as task vectors \cite{TaskVectors2024BelanecEtAl} and task arithmetic \cite{TaskArithmetic2023IlharcoEtAl} combine the adaptability of prompting with the representational control of fine-tuning, enabling efficient model editing without full retraining. 

Taken together, this suggests that future NLG systems may adopt \textit{modular, compositional control strategies}, where prompts provide rapid adaptability, fine-tuning ensures deep alignment, and decoding mechanisms enforce reliability, with hybrid approaches offering flexible trade-offs depending on the task and resource profile.

\mysection{Taxonomy of Prompting Techniques}
\vspace*{-1mm}

In this section, we present a taxonomy of prompting paradigms and how they support different NLG tasks. We distinguish between \textbf{\textit{foundational paradigms}}, which provide the basis for prompt-based control, \textbf{\textit{contextual paradigms}}, that focus on adapting prompt strategies to distinct NLG tasks and use cases, and \textbf{\textit{advanced reasoning paradigms}}, which extend prompting with search, planning, and external actions. We outline how well-suited these categories are for different NLG tasks (Table~\ref{tab:prompting_comparison}) and also present a decision flowchart in Figure~\ref{fig:flowchart} to guide practitioners in selecting an appropriate prompting strategy based on task requirements, available resources, and desired levels of control.

\subsection{Foundational Paradigms}

\myparagraph{Zero-shot Prompting~\cite{ZeroShot2019RadfordEtAl}.}
This technique relies on carefully-curated prompts to guide the LLMs in
performing specific NLG tasks, such as machine translation (MT) and story telling.  Its strength lies in the quick deployment of an idea with a relatively
low design effort.

\myparagraph{Few-shot Prompting~\cite{FewShot2020Brown2020EtAl}.}
This technique leverages the idea of in-context learning to provide 
a few input-output pairs to improve an LLM's understanding of a given 
task. Eliciting even a few high-quality examples has yielded performance
gains on different NLG tasks, such as dialogue generation and machine 
translation, steering the output towards
certain stylistic nuances.

\myparagraph{Chain-of-thought (CoT) Prompting~\cite{CoT2022WeiEtAl}.}
This technique takes its inspiration from how people decompose a complex
task into smaller sub-tasks before arriving at the final solution. Along
the same lines, this prompting paradigm provides instructions to an LLM
in such a way that encourages a step-by-step, coherent reasoning process. 
CoT can help LLMs plan outlines, sub-points and narrative arcs for NLG 
tasks such as storytelling, report generation and summarization, thereby improving global coherence.

\myparagraph{Role Prompting~\cite{RolePrompting2023KongEtAl}.}
This technique involves assigning a role to the LLM to enhance its 
understanding of the task. For example, if the model is prompted to act 
as a mathematician, it is likely to provide a correct
step-by-step explanation of a mathematical concept~\cite{PersonaSimulation2023VEtAl}. 
It also serves as an effective implicit CoT trigger, explaining its 
enhancements in reasoning capabilities. 

\subsection{Contextual Paradigms}

\myparagraph{Thread-of-thought (ThoT) Prompting~\cite{ThoT2023ZhouEtAl}.}
This technique, which draws its inspiration from human cognitive processes, 
is designed to enhance the reasoning abilities of LLM models by asking them
to select pertinent information from context comprising information from 
diverse sources, including user queries, conversation history, and external 
knowledge bases. 
ThoT has yielded substantial performance gains on tasks like conversational
question answering and dialogue system where the generation of 
contextually appropriate answer is critical.

\myparagraph{Chain-of-event (CoE) Prompting~\cite{CoE2024BaoEtAl}.}
This technique, which was proposed for summarization, consists of four steps: 
(1)~extract specific events, (2)~analyze and generalize the extracted events
in more refined and concise form, (3)~filter the generalized events to retain only those that cover most of 
the text, and (4)~integrate the events selected in step~(3) based on their 
chronological order or level of importance.

\subsection{Advanced Reasoning Paradigms}
\myparagraph{Program-of-Thought~\cite{ProgramOfThoughts2023ChenEtAl}.}
Built up on CoT, this approach represents reasoning as explicit program-like structures, improving interpretability and consistency in structured NLG tasks such as data-to-text reporting, contextual question-answering and conversational contextual question-answering tasks.

\myparagraph{Tree-of-Thoughts (ToT)~\cite{TreeofThoughts2023YaoEtAl, long2023ToT}.}
ToT expands CoT by exploring  multiple reasoning paths as a search tree, selecting the best branch for the final generation. This enhance prompting capabilities for complex tasks requiring exploration and look-ahead reasoning. 
 This structure allows language models to deliberately reason by assessing progress generated by thoughts in solving the problem. This paradigm benefits planning-heavy NLG such as story generation, report writing, and multi-step summarization.

\myparagraph{Self-Consistency Prompting~\cite{SelfConsistency2023WangEtAl}.}
Instead of relying on a single reasoning path, self-consistency samples multiple reasoning chains and aggregates the outputs. Self-Consistency utilizes a novel decoding strategy unlike the greedy one being used by CoT and consists of three key steps. The first step requires prompting the LLM using CoT, the second step samples diverse reasoning paths from LLM’s decoder and the final step involves choosing the most consistent answer across multiple reasoning paths. This improves robustness in reasoning-intensive NLG tasks such as argument generation or complex summarization.

\setlength{\tabcolsep}{2.5pt}
\begin{table*}[t]
\centering
\caption{Comparative Overview of Prompting Paradigms in NLG.\label{tab:prompting_comparison}}
\vspace*{-1mm}
{\small
\begin{tabular}{p{27mm}p{32mm}p{30mm}p{32mm}p{35mm}}
\toprule
\textbf{Prompting Paradigm} & \textbf{Primary Strategy} & \textbf{Strengths} & \textbf{Limitations} & \textbf{Typical NLG Tasks} \\
\midrule
\multicolumn{5}{c}{\textbf{Foundational Paradigms}} \\ \midrule
Zero-shot \coloredcitep{red}{ZeroShot2019RadfordEtAl} & No examples; relies on pretrained generalization & Minimal setup cost, quick deployment & Highly sensitive to prompt phrasing & Classification, QA, machine translation \\
Few-shot \coloredcitep{red}{FewShot2020Brown2020EtAl} & In-context examples demonstrate task pattern & Improves task-specific performance & Token budget grows with examples; brittle to formatting & Summarization, dialogue generation, structured text \\
Chain-of-Thought \coloredcitep{red}{CoT2022WeiEtAl} & Step-by-step reasoning in prompts & Improves reasoning transparency and coherence & Verbose; task-dependent gains & Explainable generation, QA, structured reports \\
Role Prompting \coloredcitep{red}{RolePrompting2023KongEtAl} & Assigns persona or role to guide behavior & Boosts creativity, diversity, and framing & Relies on accurate persona crafting & Dialogue agents, creative writing, tutoring \\ \midrule
\multicolumn{5}{c}{\textbf{Contextual Paradigms}} \\ \midrule
Thread-of-Thought \coloredcitep{red}{ThoT2023ZhouEtAl} & Maintains discourse flow across turns & Enhances context tracking in long dialogues & Hard to automate; needs clear structure & Conversational QA, data-to-text NLG \\
Chain-of-Event \coloredcitep{red}{CoE2024BaoEtAl} & Extracts and compresses event chains & Improves summary coherence and fluency & Narrow scope; task-specific design & Multi-document summarization \\ \midrule
\multicolumn{5}{c}{\textbf{Advanced Reasoning Paradigms}} \\ \midrule
Program-of-Thought \coloredcitep{red}{ProgramOfThoughts2023ChenEtAl} & Converts reasoning into program-like steps & Improves interpretability and structured reasoning & Requires task formalization & Data-to-text generation, structured reporting \\
Tree-of-Thoughts \coloredcitep{red}{TreeofThoughts2023YaoEtAl,long2023ToT} & Explores reasoning as a search tree & Supports planning and complex reasoning & Computationally expensive & Story generation, multi-step summarization, planning-heavy NLG \\
Self-Consistency \coloredcitep{red}{SelfConsistency2023WangEtAl} & Aggregates multiple reasoning chains & Improves robustness and factual reliability & Costly; may over-generate redundant outputs & Argument generation, complex summarization \\

\bottomrule
\end{tabular}
}
\vspace*{-2mm}
\end{table*}

\begin{comment}

\begin{table}[t]
\centering
\caption{Prompting Paradigms and Their Relevance to NLG Tasks}
\label{tab:prompting_paradigms_nlg_tasks}
\renewcommand{\arraystretch}{1.15}
\small
\begin{tabularx}{\linewidth}{|>{\raggedright\arraybackslash}p{3cm}|X|}
\hline
\textbf{Prompting Paradigm} & \textbf{Relevant NLG Tasks} \\
\hline
Zero-shot Prompting \cite{ZeroShot2019RadfordEtAl} & Machine translation, paraphrasing, question generation  \\
\hline
Few-shot Prompting \cite{FewShot2020Brown2020EtAl} & Summarization, translation, structured generation \\
\hline
Chain-of-thought Prompting \cite{CoT2022WeiEtAl} & Reasoning-based question answering, arithmetic explanation generation, data-to-text generation\\
\hline
Thread-of-thought Prompting [\cite{ThoT2023ZhouEtAl}] & Conversational QA, long-context scenarios \\
\hline
Chain-of-event Prompting [\cite{CoE2024BaoEtAl}] & Multi-document summarization \\
\hline
Role-play Prompting \cite{RolePrompting2023KongEtAl} & Dialogue agents, creative writing \\
\hline
\end{tabularx}
\end{table}

\end{comment}

\mysection{Prompting for Controlled NLG}
\vspace*{-1mm}
Despite the LLMs ability to generate fluent and grammatically 
sound texts, other control dimensions are important in real-world
applications, \eg style, content, length and structure, which 
might not be achievable if we allow LLMs to generate text 
freely~\cite{PlugPlay2024AjwaniEtAl}. Controlling model outputs to 
fit a set of constraints is the purview of controllable text 
generation~\cite{TopicControl2023LuEtAl}, and prompting serves as 
a \textit{`control lever'} for steering the outputs of LLMs,
without requiring extensive parameter fine-tuning. 

\mysubsection{Content Control: Topical Constraints and Lexical Anchoring}

The limits of generative LLMs are unclear, yet from research perspective,
we must be able to determine what makes them successful and what causes 
them to fail~\cite{TopicControl2023LuEtAl}. 
Carefully crafted prompts can enforce topical constraints by explicitly 
specifying the domain or keywords to be included in the output. For example,
``Generate a short explanation about the challenges blind and low vision individuals face in accessing data visualizations in educational settings'' can anchor the generation 
of text around the relevant topic and subtopics. 
Similarly, lexical anchoring, 
which refers to specifying important information to the LLM before reaching a decision,
can guide the model to generate domain-specific information without 
re-training~\cite{PromptAnchoring2024TianEtAl}. For example, prompting with 
``Write a paragraph about the challenges in web accessibility for blind users. Make sure to include the terms \textit{tactile graphics}, \textit{screen reader}, and \textit{access barrier}"  anchors
the model in accessibility terminology and context.
These approaches are useful for NLG tasks such as
question answering~\cite{GeneratedQA2021ZhuEtAl}, story telling~\cite{StoryGeneration2018FanEtAl}, and summarization \cite{SalientFeatures2024XxuEtAl} etc. 
\mysubsection{Structure Control: Length and Discourse Organization }
Users often expect generated texts to fall within a specific length 
range, making length controlled generation an important topic~\cite{LengthControl2024JieEtAl2024}. 
For pre-trained language models, the most widely applied technique for 
length control is prompt-based fine-tuning (\eg `summarize in two sentences' or `provide a bullet-point list of advantages of living in coastal areas)~\cite{Pretrain2023LiuEtAl}. 
In addition, prompts can also guide an LLM to simulate the discourse 
structure of human-written text~\cite{Discourse2022GhazvinineEtAl}. 

\mysubsection{Style Control: Formality, Emotion, and Tone}
Prompting can effectively condition the formality level (`write in a 
formal academic style' vs. `explain in a casual, friendly tone'), 
emotional coloring (`write a comforting message to someone anxious 
about exams'), and persona-based tone ('as a supervisor, advise on 
thesis writing'). Drawing its inspiration from the impact of language
on human performance, EmotionPrompt~\cite{EmotionPrompt2023LiEtAl} 
utilizes emotional cues to help improve an LLM's emotional appeal. 
Compared to style transfer methods requiring paired data or explicit 
attribute modeling, prompting offers a flexible 
and low-resource alternative \cite{StyleTransfer2025YangEtAl}. In addition to written output, the emotional prompts may be used to control and guide the emotional tone of synthesized speech~\cite{EmotionControl2024BottEtAl}.

While prompting is effective for controlling individual stylistic dimensions such as formality or tone, achieving \textit{fine-grained multi-attribute control} (e.g., simultaneously adjusting formality, sentiment, and persona) remains a significant challenge. Prompt formulations often lead to conflicts between attributes, resulting in outputs that privilege one dimension over another. Recent decoding-time strategies, such as contrastive decoding \cite{ShiContrastiveDecoding2023}, highlight the importance of leveraging context-aware constraints to balance competing stylistic goals. However, extending such approaches to multi-attribute style control via prompting alone remains underexplored and may require hybrid solutions that combine manual prompt design, optimization-based methods, and decoding-level interventions.

\mysection{Evaluation and Prompt Robustness}
To systematically understand the impact of prompt engineering, we need to examine how different evaluation metrics capture output quality. Unlike standard NLG, prompted NLG introduces unique challenges: performance is often more sensitive to phrasing, reasoning depth, and context handling. As such, traditional metrics may not fully reflect the effectiveness of prompt-based control.

Both \textbf{\textit{human-centered}} and \textbf{\textit{automatic}} metrics have traditionally been used to assess NLG quality, each with distinct advantages and limitations. More recently, a third paradigm, known as \textbf{\textit{LLM-as-a-Judge}}, has emerged, leveraging LLMs themselves as evaluators to complement human and automatic methods.

\paragraph{Human Evaluation.} Human evaluation relies on expert or crowd annotators to assess dimensions that are difficult to quantify automatically, such as coherence, fluency, factuality, and stylistic appropriateness \cite{NeuralTextDegeneration2020HoltzmanEtAl}. It is increasingly used in abstract tasks like summarization and creative writing \cite{Summarization2020StiennonEtAl,TreeofThoughts2023YaoEtAl,DualCritiquePrompting2024WangEtAl}. However, human evaluation is costly, time-consuming, and may suffer from annotator bias or inconsistency, making large-scale benchmarking difficult.

\paragraph{Automatic Evaluation.} Automatic metrics are fast and scalable, providing quantitative assessments of different prompting strategies. Classical overlap-based metrics such as BLEU \cite{BLEU2002PapineniEtAl}, ROUGE \cite{ROUGE2004LinEtAl}, and METEOR \cite{METEOR2005BanerjeeEtAl} remain widely used, while embedding-based metrics such as BERTScore \cite{BERTScore2019ZhangEtAl} and MoverScore offer higher semantic alignment. More recent approaches include \textit{contrastive decoding} methods \cite{ShiContrastiveDecoding2023}, which evaluate preference between candidate generations, and specialized metrics for safety, such as toxicity and bias scores \cite{ToxicityScore2020GehmanEtAl}. Despite their efficiency, automatic metrics often fail to capture nuanced qualities like creativity, discourse structure, or reasoning validity \cite{MetricsEvaluationSurvey2022SaiEtAl}.

\paragraph{LLM-as-a-Judge.} A recent trend is to use LLMs themselves as evaluators. Here, an LLM is prompted to rate outputs according to quality dimensions such as coherence, faithfulness, or style. Studies show that GPT-4 and similar models correlate well with expert human judgments, enabling scalable and cost-efficient evaluation \cite{Zheng2023LLMJudge}. However, this approach raises concerns that LLM evaluators may inherit the biases of the model being judged, may be vulnerable to prompt hacks, and risk circularity when the same model family is used for both generation and evaluation \cite{SurverLLMAsJudge2024GuEtAl}. Developing reliable protocols for “LLM-as-a-judge” remains an open challenge.

\paragraph{Discussion.}
Evaluating prompted NLG thus remains difficult for several reasons: 
\begin{itemize}
    \item Sensitivity to minor prompt variations complicates reproducibility of evaluation results \cite{SclarEtAl2023Quantifying}; 
    \item Existing metrics underrepresent attributes like reasoning faithfulness and contextual coherence \cite{LiEtAl2024LlmsAsAJudge}; 
    \item LLM-based judges, while promising, raise concerns about bias propagation and lack of interpretability \cite{NovikovaEtAl2025Consistency}. 
\end{itemize}
A comprehensive evaluation framework should therefore triangulate across human, automatic, and LLM-based metrics to capture the full spectrum of quality \cite{EvaluLLM2024PanEtAl}, especially as prompting becomes central to output-level control.

 \mysubsection{Prompt Sensitivity and Brittleness, and Optimization}
LLM-generated outputs can be highly sensitive  and variable to the prompt phrasing and structure, with performance often differing markedly across models and tasks based on these nuances \cite{Prosa2024ZhuoEtAl,PosixC2024ChatterjeeEtAl,ButterflyEffect2024SalinasEtAl}. This sensitivity manifests in two formats: i) \textit{local sensitivity} refers to the small lexical changes such as synonyms or reordering of instructions may lead to noticeable difference in tone, quality and relevance of the generated response. ii). \textit{global brittleness} refers to changes in prompt length or specificity can lead to failures in adhering to the intended task, indicating a lack of robustness in prompt-based control compared to more structured fine-tuning methods.

To mitigate these issues, the field of \textbf{\textit{prompt optimization}} has emerged. Instead of relying solely on manual trial-and-error, optimization methods learn or refine prompts in a systematic manner. 
\textit{Prompt tuning} \cite{PromptTuning2021Lester} introduces soft prompt vectors learned while keeping model parameters frozen, offering consistency across tasks. 
\textit{Prefix tuning} \cite{PrefixTuning2021Li} similarly prepends trainable vectors to the input sequence, enabling parameter-efficient adaptation. 
Other approaches such as \textit{task vectors} and \textit{task arithmetic} \cite{TaskVectors2024BelanecEtAl, TaskArithmetic2023IlharcoEtAl} allow compositional editing of model behavior by adding or subtracting task-specific directions in embedding space. 
Automated prompt search methods, including gradient-based and reinforcement-learning approaches \cite{AutoPrompt2021,BlackBoxPromptOpt2023}, further extend this line by discovering effective discrete prompts without human intervention.

While these optimization strategies enhance robustness and portability, they trade off interpretability and creative flexibility \cite{PatelEtAl2025Optimization}. Manual prompt design remains attractive for low-resource, domain-specific tasks, whereas optimization-based methods provide scalability, reproducibility, and systematic control. 
A comprehensive framework for when to employ manual design, optimization, or hybrid approaches remains an important open research question.

\mysection{Emerging Trends and Open Challenges}
The field of prompt engineering is evolving through various key trends. \textit{Prompting with retrieval augmented generation} (RAG) \cite{RAG2020LewisEtAl} enforce factual grounding and reduce hallucinations by pairing text generation with retrieval. \textit{Prompt engineering as programming} \cite{SPML2024SharmaEtAl} is capturing attentions via prompt DSL, a domain-specific language for refining prompts and monitoring the inputs to the LLM-based chatbots. More recently, the concept of \textit{context engineering }\cite{ContextEngineering2025MeiEtAl} has been introduced, extending prompt design into a broader process of selecting, organizing, and weighting relevant contextual information before it is provided to the model.
This approach views prompts not as standalone inputs but as components of a larger pipeline that integrates retrieval, augmentation, and orchestration strategies. In doing so, it connects prompt engineering with broader system-level design of LLM applications. Another emerging trend is the application of prompt engineering in \textit{multilingual NLG} \cite{VatsalEtAl2025Multilingual}.
Recent studies shows that prompts can enable cross-lingual transfer and adaptation, allowing LLMs to generalize across diverse linguistic and cultural contexts.
However, multilingual prompting also highlights new opportunities for advancing context-aware and culturally grounded design, where lexical choices, stylistic framing, and contextual cues are adapted to language-specific conventions rather than translated directly from English.

Despite its promise, prompt engineering faces several persistent challenges that limit its reliability and generalizability.

\mysubsection{Prompt Brittleness}
LLM outputs are highly sensitive to small variations in prompt wording, ordering, or formatting. Even minor lexical substitutions can produce drastically different generations, leading to instability in downstream applications \cite{ButterflyEffect2024SalinasEtAl,PosixC2024ChatterjeeEtAl,Prosa2024ZhuoEtAl}. This brittleness raises concerns for reproducibility and makes it difficult to establish robust prompt design principles.

\mysubsection{Bias Amplification}
Prompts can inadvertently reinforce or even amplify social biases embedded in pre-trained LLMs. For example, toxicity benchmarks have shown that prompts conditioned on specific demographics can trigger harmful stereotypes \cite{ToxicityScore2020GehmanEtAl}. While lexical anchoring and prompt tuning may mitigate these risks, bias amplification remains a core obstacle for deploying prompt-engineered systems responsibly.

\mysubsection{Factual Accuracy and Hallucination}
Although prompt engineering can steer style and content, it often fails to ensure factual accuracy. LLMs are prone to hallucinating details or presenting plausible-sounding but incorrect information. Retrieval-augmented prompting \cite{RAG2020LewisEtAl} has shown promise, but ensuring factual reliability across domains remains unresolved.

\mysubsection{Evaluation Gaps}
Evaluating prompts is challenging. Human evaluations provide nuanced insights but are costly and inconsistent, while automatic metrics (e.g., BLEU, ROUGE, BERTScore) capture only surface similarity. Recent trends in using LLMs as judges offer scalability but introduce new risks of model bias \cite{MetricsEvaluationSurvey2022SaiEtAl,EvaluationTextGeneration2020CelikyilmazEtAl}. Without standardized benchmarks, it is difficult to compare techniques fairly.

\mysubsection{Generalization and Portability}
Prompts designed for one domain, language, or model architecture often fail to transfer effectively \cite{Transferability2021PerezEtAl}. This lack of portability is particularly problematic in multilingual and cross-domain NLG, where prompts must adapt to diverse contexts without extensive re-engineering.

\mysection{Toward a Systematic Framework for Prompt Engineering}

\begin{table*}[h
t!]
\centering
\small
\caption{Core dimensions of a systematic framework for prompt engineering.}
\begin{tabular}{@{}p{2cm}p{4cm}p{6cm}p{3.5cm}@{}}
\toprule
\textbf{Dimension} & \textbf{Core Practices} & \textbf{Illustrative Methods / References} & \textbf{Trade-offs} \\ 
\midrule
\textbf{Design} & 
Abstraction over specificity; modular, reusable prompt templates & 
Role-based and task-agnostic prompting~\coloredcitep{red}{PromptEng2024ShubhamEtAl}; multilingual and culturally neutral phrasing~\coloredcitep{red} {VatsalEtAl2025Multilingual} & 
May reduce creativity or cultural nuance \\ 
\midrule
\textbf{Optimization} & 
Lightweight adaptation; search-based or continuous tuning & 
Discrete prompt search (AutoPrompt)~\coloredcitep{red}{AutoPrompt2021}; Prefix Tuning~\coloredcitep{red}{PrefixTuning2021Li}; Prompt Tuning~\coloredcitep{red}{PromptTuning2021Lester}; Contrastive Decoding~\coloredcitep{red}{ShiContrastiveDecoding2023} & 
Reduced interpretability; limited creative flexibility \\ 
\midrule
\textbf{Evaluation} & 
Systematic stress-testing across domains and languages & 
BLEU~\coloredcitep{red}{BLEU2002PapineniEtAl}, ROUGE~\coloredcitep{red}{ROUGE2004LinEtAl}, BERTScore~\cite{BERTScore2019ZhangEtAl}; Bias and Toxicity scores~\coloredcitep{red}{ToxicityScore2020GehmanEtAl}; LLM-as-a-Judge~\coloredcitep{red}{Zheng2023LLMJudge} & 
Metric bias; reproducibility and transparency challenges \\ 
\bottomrule
\end{tabular}
\label{tab:framework_best_practices}
\end{table*}

While existing research highlights the fragility and limited portability of prompts across domains, languages, and model architectures \cite{Transferability2021PerezEtAl}, the field still lacks a unified framework that connects design principles, optimization strategies, and evaluation practices. 
Establishing such a framework is critical for advancing prompt engineering from an empirical craft to a principled science of input-level control.

We propose that progress in prompt engineering can be structured around three interdependent dimensions namely:\textbf{\textit{ i. design}}, \textbf{\textit{ii. optimization}}, and \textbf{\textit{iii. evaluation}}. 
Each dimension captures a complementary aspect of how prompts are constructed, adapted, and validated. 
\textit{Design} focuses on abstraction and reusability; \textit{optimization} emphasizes robustness and scalability; and \textit{evaluation} ensures reliability, fairness, and transferability across tasks and languages. 
Table~\ref{tab:framework_best_practices} summarizes these dimensions, the methods that operationalize them, and the trade-offs that arise when balancing interpretability, flexibility, and control.

\subsection{Design Principles}
Prompt design should prioritize abstraction over specificity. Reusable templates, role-based prompting, and culturally neutral phrasing facilitate cross-domain and multilingual portability \cite{PromptEng2024ShubhamEtAl,VatsalEtAl2025Multilingual}. At the same time, excessive abstraction may reduce expressiveness or limit stylistic nuance, underscoring the need for modular design that retains both flexibility and generality.

\subsection{Optimization}
Portability often depends on lightweight adaptation mechanisms that refine prompts without retraining the entire model. 
Methods such as discrete prompt search~\cite{AutoPrompt2021}, Prefix and Prompt Tuning~\cite{PrefixTuning2021Li,PromptTuning2021Lester}, and decoding-level interventions like contrastive decoding~\cite{ShiContrastiveDecoding2023} exemplify scalable solutions for aligning prompts across architectures. 
Yet these optimization-based strategies frequently trade interpretability and creative flexibility for robustness, raising open questions about how to balance manual and automated prompt design \cite{PatelEtAl2025Optimization}.

\subsection{Evaluation}
A systematic framework also requires rigorous stress-testing across linguistic, cultural, and domain boundaries. Complementary metrics, BLEU, ROUGE, BERTScore, quantify task performance, while safety-oriented scores such as toxicity and bias~\cite{ToxicityScore2020GehmanEtAl} and emerging model-based evaluators (LLM-as-a-Judge) ~\cite{Zheng2023LLMJudge,SurverLLMAsJudge2024GuEtAl} assess quality and fairness. Ensuring the transparency and reproducibility of such evaluations remains a central research priority.

\mysection{Conclusion}
Prompting has evolved into a core technique for LLM-driven NLG, 
enabling controlled, efficient generation without retraining. However, prompt design often involves repetitive and time-consuming debugging, as small phrasing changes can lead to unpredictable outputs.

To advance the field, prompt engineering should be formalized
with frameworks, benchmarks and theory that support robust,
scalable, and reusable prompt design for the next generation of NLG 
systems. There is also a pressing need for evaluation metrics that more accurately reflect the effectiveness and control capabilities of prompts.

\mysection{Limitations}
\vspace*{-1mm}
Despite offering a focused overview of prompt engineering for NLG, this survey has several limitations. Findings discussed are largely conceptual and may not generalize across domains, languages, or LLM architectures without further empirical validation. While we highlight issues such as prompt sensitivity and evaluation gaps, practical mitigation strategies and theoretical modeling could not be explored in depth. Future work is needed to extend the taxonomy, assess generalizability across multilingual settings, and ground prompt engineering in more rigorous empirical and theoretical frameworks.

\bibliography{main}

@article{PromptEng2024ShubhamEtAl,
  author       = {Shubham Vatsal and Harsh Dubey},
  title        = {A Survey of Prompt Engineering Methods in Large Language Models for
                  Different {NLP} Tasks},
  journal      = {CoRR},
  volume       = {abs/2407.12994},
  year         = {2024},
  url          = {https://doi.org/10.48550/arXiv.2407.12994},
  doi          = {10.48550/ARXIV.2407.12994},
}

@article{Pretrain2023LiuEtAl,
  title={Pre-train, prompt, and predict: A systematic survey of prompting methods in natural language processing},
  author={Liu, Pengfei and Yuan, Weizhe and Fu, Jinlan and Jiang, Zhengbao and Hayashi, Hiroaki and Neubig, Graham},
  journal={ACM computing surveys},
  volume={55},
  number={9},
  pages={1--35},
  year={2023},
}

@article{Unleashing2025ChenEtAl,
  title={Unleashing the potential of prompt engineering for large language models},
  author={Chen, Banghao and Zhang, Zhaofeng and Langren{\'e}, Nicolas and Zhu, Shengxin},
  journal={Patterns},
  year={2025},
  publisher={Elsevier}
}

@article{NLG2018GattEtAl,
  author       = {Albert Gatt and
                  Emiel Krahmer},
  title        = {Survey of the State of the Art in Natural Language Generation: Core
                  tasks, applications and evaluation},
  journal      = {J. Artif. Intell. Res.},
  volume       = {61},
  pages        = {65--170},
  year         = {2018},
  doi          = {10.1613/JAIR.5477},

}

@article{SequentialNLG2023PandeyEtAl,
  author       = {Abhishek Kumar Pandey and
                  Sanjiban Sekhar Roy},
  title        = {Natural Language Generation Using Sequential Models: {A} Survey},
  journal      = {Neural Process. Lett.},
  volume       = {55},
  number       = {6},
  pages        = {7709--7742},
  year         = {2023},
  doi          = {10.1007/S11063-023-11281-6},
}

@article{Parameter2023XuEtAl,
  title={Parameter-efficient fine-tuning methods for pretrained language models: A critical review and assessment},
  author={Xu, Lingling and Xie, Haoran and Qin, Si-Zhao Joe and Tao, Xiaohui and Wang, Fu Lee},
  journal={arXiv preprint arXiv:2312.12148},
  year={2023}
}

@article{ZeroShot2019RadfordEtAl,
  title={Language models are unsupervised multitask learners},
  author={Radford, Alec and Wu, Jeffrey and Child, Rewon and Luan, David and Amodei, Dario and Sutskever, Ilya and others},
  journal={OpenAI blog},
  volume={1},
  number={8},
  pages={9},
  year={2019}
}

@article{FewShot2020Brown2020EtAl,
  title={Language models are few-shot learners},
  author={Brown, Tom and Mann, Benjamin and Ryder, Nick and Subbiah, Melanie and Kaplan, Jared D and Dhariwal, Prafulla and Neelakantan, Arvind and Shyam, Pranav and Sastry, Girish and Askell, Amanda and others},
  journal={Advances in neural information processing systems},
  volume={33},
  pages={1877--1901},
  year={2020}
}

@article{CoT2022WeiEtAl,
  title={Chain-of-thought prompting elicits reasoning in large language models},
  author={Wei, Jason and Wang, Xuezhi and Schuurmans, Dale and Bosma, Maarten and Xia, Fei and Chi, Ed and Le, Quoc V and Zhou, Denny and others},
  journal={Advances in neural information processing systems},
  volume={35},
  pages={24824--24837},
  year={2022}
}

@article{RolePrompting2023KongEtAl,
  title={Better zero-shot reasoning with role-play prompting},
  author={Kong, Aobo and Zhao, Shiwan and Chen, Hao and Li, Qicheng and Qin, Yong and Sun, Ruiqi and Zhou, Xin and Wang, Enzhi and Dong, Xiaohang},
  journal={arXiv preprint arXiv:2308.07702},
  year={2023}
}

@article{ThoT2023ZhouEtAl,
  title={Thread of thought unraveling chaotic contexts},
  author={Zhou, Yucheng and Geng, Xiubo and Shen, Tao and Tao, Chongyang and Long, Guodong and Lou, Jian-Guang and Shen, Jianbing},
  journal={arXiv preprint arXiv:2311.08734},
  year={2023}
}

@article{CoE2024BaoEtAl,
  title={Chain-of-event prompting for multi-document summarization by large language models},
  author={Bao, Songlin and Li, Tiantian and Cao, Bin},
  journal={International Journal of Web Information Systems},
  number={ahead-of-print},
  year={2024},
}

@article{PlugPlay2024AjwaniEtAl,
  title={Plug and Play with Prompts: A Prompt Tuning Approach for Controlling Text Generation},
  author={Ajwani, Rohan Deepak and Zhu, Zining and Rose, Jonathan and Rudzicz, Frank},
  journal={arXiv preprint arXiv:2404.05143},
  year={2024}
}

@article{LengthControl2024JieEtAl2024,
  title={Prompt-based length controlled generation with multiple control types},
  author={Jie, Renlong and Meng, Xiaojun and Shang, Lifeng and Jiang, Xin and Liu, Qun},
  journal={arXiv preprint arXiv:2406.10278},
  year={2024}
}

@inproceedings{Discourse2022GhazvinineEtAl,
  author       = {Marjan Ghazvininejad and
                  Vladimir Karpukhin and
                  Vera Gor and
                  Asli Celikyilmaz},
  title        = {Discourse-Aware Soft Prompting for Text Generation},
  booktitle    = {Proceedings of the 2022 Conference on Empirical Methods in Natural
                  Language Processing, {EMNLP} 2022, Abu Dhabi, United Arab Emirates,
                  December 7-11, 2022},
  pages        = {4570--4589},
  year         = {2022},
  doi          = {10.18653/V1/2022.EMNLP-MAIN.303},

}

@article{EmotionPrompt2023LiEtAl,
  title={Large language models understand and can be enhanced by emotional stimuli},
  author={Li, Cheng and Wang, Jindong and Zhang, Yixuan and Zhu, Kaijie and Hou, Wenxin and Lian, Jianxun and Luo, Fang and Yang, Qiang and Xie, Xing},
  journal={arXiv preprint arXiv:2307.11760},
  year={2023}
}

@article{EmotionControl2024BottEtAl,
  title={Controlling emotion in text-to-speech with natural language prompts},
  author={Bott, Thomas and Lux, Florian and Vu, Ngoc Thang},
  journal={arXiv preprint arXiv:2406.06406},
  year={2024}
}

@inproceedings{TopicControl2023LuEtAl,
    title = "Bounding the Capabilities of Large Language Models in Open Text Generation with Prompt Constraints",
    author = "Lu, Albert  and
      Zhang, Hongxin  and
      Zhang, Yanzhe  and
      Wang, Xuezhi  and
      Yang, Diyi",
    booktitle = "Findings of the Association for Computational Linguistics: EACL 2023",
    month = may,
    year = "2023",
    doi = "10.18653/v1/2023.findings-eacl.148",
    pages = "1982--2008",
   
}

@inproceedings{StoryGeneration2018FanEtAl,
  author       = {Angela Fan and
                  Mike Lewis and
                  Yann N. Dauphin},
  title        = {Hierarchical Neural Story Generation},
  booktitle    = {Proceedings of the 56th Annual Meeting of the Association for Computational
                  Linguistics, {ACL} 2018, Melbourne, Australia, July 15-20, 2018, Volume
                  1: Long Papers},
  pages        = {889--898},
  publisher    = {Association for Computational Linguistics},
  year         = {2018},
  doi          = {10.18653/V1/P18-1082},
}

@article{GeneratedQA2021ZhuEtAl,
  author       = {Fengbin Zhu and
                  Wenqiang Lei and
                  Chao Wang and
                  Jianming Zheng and
                  Soujanya Poria and
                  Tat{-}Seng Chua},
  title        = {Retrieving and Reading: {A} Comprehensive Survey on Open-domain Question
                  Answering},
  journal      = {CoRR},
  volume       = {abs/2101.00774},
  year         = {2021},
  url          = {https://arxiv.org/abs/2101.00774},
  eprinttype    = {arXiv},
  eprint       = {2101.00774},
}

@article{PromptAnchoring2024TianEtAl,
  author       = {Yuan Tian and
                  Tianyi Zhang},
  title        = {Selective Prompt Anchoring for Code Generation},
  journal      = {CoRR},
  volume       = {abs/2408.09121},
  year         = {2024},
  url          = {https://doi.org/10.48550/arXiv.2408.09121},
  doi          = {10.48550/ARXIV.2408.09121},
  eprinttype    = {arXiv},
  eprint       = {2408.09121},
}

@inproceedings{PromptvsFineTuning2025ShinEtAl,
  title={Prompt Engineering or Fine-Tuning: An Empirical Assessment of LLMs for Code},
  author={Shin, Jiho and Tang, Clark and Mohati, Tahmineh and Nayebi, Maleknaz and Wang, Song and Hemmati, Hadi},
  booktitle={2025 IEEE/ACM 22nd International Conference on Mining Software Repositories (MSR)},
  pages={490--502},
  year={2025},
  organization={IEEE}
}

@article{PersonaSimulation2023VEtAl,
  title={Guided scenarios with simulated expert personae: a remarkable strategy to perform cognitive work},
  author={Van Buren, David},
  journal={arXiv preprint arXiv:2306.03104},
  year={2023}
}

@inproceedings{EvaluationMetric2005StentEtAl,
author = {Stent, Amanda and Marge, Matthew and Singhai, Mohit},
title = {Evaluating evaluation methods for generation in the presence of variation},
year = {2005},
isbn = {3540245235},
url = {https://doi.org/10.1007/978-3-540-30586-6_38},
doi = {10.1007/978-3-540-30586-6_38},
pages = {341–351},

}

@inproceedings{NeuralTextDegeneration2020HoltzmanEtAl,
  author       = {Ari Holtzman and
                  Jan Buys and
                  Li Du and
                  Maxwell Forbes and
                  Yejin Choi},
  title        = {The Curious Case of Neural Text Degeneration},
  booktitle    = {8th International Conference on Learning Representations, {ICLR} 2020,
                  Addis Ababa, Ethiopia, April 26-30, 2020},
  publisher    = {OpenReview.net},
  year         = {2020},
}

@inproceedings{Summarization2020StiennonEtAl,
author = {Stiennon, Nisan and Ouyang, Long and Wu, Jeff and Ziegler, Daniel M. and Lowe, Ryan and Voss, Chelsea and Radford, Alec and Amodei, Dario and Christiano, Paul},
title = {Learning to summarize from human feedback},
year = {2020},
booktitle = {Proceedings of the 34th International Conference on Neural Information Processing Systems},

}

@inproceedings{TreeofThoughts2023YaoEtAl,
author = {Yao, Shunyu and Yu, Dian and Zhao, Jeffrey and Shafran, Izhak and Griffiths, Thomas L. and Cao, Yuan and Narasimhan, Karthik},
title = {Tree of thoughts: deliberate problem solving with large language models},
year = {2023},
booktitle = {Proceedings of the 37th International Conference on Neural Information Processing Systems},}

@inproceedings{DualCritiquePrompting2024WangEtAl,
    title = "Enhancing Large Language Models Against Inductive Instructions with Dual-critique Prompting",
    author = "Wang, Rui  and
      Wang, Hongru  and
      Mi, Fei  and
      Xue, Boyang  and
      Chen, Yi  and
      Wong, Kam-Fai  and
      Xu, Ruifeng",
    booktitle = "Proceedings of the 2024 Conference of the North American Chapter of the Association for Computational Linguistics: Human Language Technologies (Volume 1: Long Papers)",
    year = "2024",
    pages = "5345--5363",
}

@inproceedings{BLEU2002PapineniEtAl,
author = {Papineni, Kishore and Roukos, Salim and Ward, Todd and Zhu, Wei-Jing},
title = {BLEU: a method for automatic evaluation of machine translation},
year = {2002},
booktitle = {Proceedings of the 40th Annual Meeting on Association for Computational Linguistics},
pages = {311–318},

}

@inproceedings{ROUGE2004LinEtAl,
    title = "{ROUGE}: A Package for Automatic Evaluation of Summaries",
    author = "Lin, Chin-Yew",
    booktitle = "Text Summarization Branches Out",
    year = "2004",
    pages = "74--81"
}

@inproceedings{METEOR2005BanerjeeEtAl,
    title = "{METEOR}: An Automatic Metric for {MT} Evaluation with Improved Correlation with Human Judgments",
    author = "Banerjee, Satanjeev  and
      Lavie, Alon",
    booktitle = "Proceedings of the {ACL} Workshop on Intrinsic and Extrinsic Evaluation Measures for Machine Translation and/or Summarization",
    year = "2005",
    pages = "65--72"
}

@article{BERTScore2019ZhangEtAl,
  title={Bertscore: Evaluating text generation with bert},
  author={Zhang, Tianyi and Kishore, Varsha and Wu, Felix and Weinberger, Kilian Q and Artzi, Yoav},
  journal={arXiv preprint arXiv:1904.09675},
  year={2019}
}

@article{MetricsEvaluationSurvey2022SaiEtAl,
author = {Sai, Ananya B. and Mohankumar, Akash Kumar and Khapra, Mitesh M.},
title = {A Survey of Evaluation Metrics Used for NLG Systems},
year = {2022},
publisher = {Association for Computing Machinery},
address = {New York, NY, USA},
volume = {55},
number = {2},
issn = {0360-0300},
journal = {ACM Computing Survey},

}

@inproceedings{ToxicityScore2020GehmanEtAl,
    title = "{R}eal{T}oxicity{P}rompts: Evaluating Neural Toxic Degeneration in Language Models",
    author = "Gehman, Samuel  and
      Gururangan, Suchin  and
      Sap, Maarten  and
      Choi, Yejin  and
      Smith, Noah A.",
    booktitle = "Findings of the Association for Computational Linguistics: EMNLP 2020",
    year = "2020",
    pages = "3356--3369",
}

@inproceedings{Prosa2024ZhuoEtAl,
    title = "{P}ro{SA}: Assessing and Understanding the Prompt Sensitivity of {LLM}s",
    author = "Zhuo, Jingming  and
      Zhang, Songyang  and
      Fang, Xinyu  and
      Duan, Haodong  and
      Lin, Dahua  and
      Chen, Kai",
    booktitle = "Findings of the Association for Computational Linguistics: EMNLP 2024",
    year = "2024",
    pages = "1950--1976",
}

@inproceedings{PosixC2024ChatterjeeEtAl,
    title = "{POSIX}: A Prompt Sensitivity Index For Large Language Models",
    author = "Chatterjee, Anwoy  and
      Renduchintala, H S V N S Kowndinya  and
      Bhatia, Sumit  and
      Chakraborty, Tanmoy",
    booktitle = "Findings of the Association for Computational Linguistics: EMNLP 2024",
    year = "2024",
    doi = "10.18653/v1/2024.findings-emnlp.852",
    pages = "14550--14565",

}

@inproceedings{ButterflyEffect2024SalinasEtAl,
    title = "The Butterfly Effect of Altering Prompts: How Small Changes and Jailbreaks Affect Large Language Model Performance",
    author = "Salinas, Abel  and
      Morstatter, Fred",
    booktitle = "Findings of the Association for Computational Linguistics: ACL 2024",
    year = "2024",
    doi = "10.18653/v1/2024.findings-acl.275",
    pages = "4629--4651",
}

@inproceedings{PromptTuning2021Lester,
    title = "The Power of Scale for Parameter-Efficient Prompt Tuning",
    author = "Lester, Brian  and
      Al-Rfou, Rami  and
      Constant, Noah",
    booktitle = "Proceedings of the 2021 Conference on Empirical Methods in Natural Language Processing",
    year = "2021",
    pages = "3045--3059",

}

@article{RAG2020LewisEtAl,
  title={Retrieval-augmented generation for knowledge-intensive nlp tasks},
  author={Lewis, Patrick and Perez, Ethan and Piktus, Aleksandra and Petroni, Fabio and Karpukhin, Vladimir and Goyal, Naman and K{\"u}ttler, Heinrich and Lewis, Mike and Yih, Wen-tau and Rockt{\"a}schel, Tim and others},
  journal={Advances in neural information processing systems},
  volume={33},
  pages={9459--9474},
  year={2020}
}

@article{SPML2024SharmaEtAl,
  title={{SPML}: A dsl for defending language models against prompt attacks},
  author={Sharma, Reshabh K and Gupta, Vinayak and Grossman, Dan},
  journal={arXiv preprint arXiv:2402.11755},
  year={2024}
}

@article{Transferability2021PerezEtAl,
  title={True few-shot learning with language models},
  author={Perez, Ethan and Kiela, Douwe and Cho, Kyunghyun},
  journal={Advances in neural information processing systems},
  volume={34},
  pages={11054--11070},
  year={2021}
}

@article{NLG2023DongEtAl,
author = {Dong, Chenhe and Li, Yinghui and Gong, Haifan and Chen, Miaoxin and Li, Junxin and Shen, Ying and Yang, Min},
title = {A Survey of Natural Language Generation},
year = {2022},
issue_date = {August 2023},
publisher = {Association for Computing Machinery},
address = {New York, NY, USA},
volume = {55},
number = {8},
issn = {0360-0300},
journal = {ACM Computing Survey},

}

@article{NLGSequentialArchitectures2023PandeyEtAl,
  title={Natural language generation using sequential models: A survey},
  author={Pandey, Abhishek Kumar and Roy, Sanjiban Sekhar},
  journal={Neural Processing Letters},
  volume={55},
  number={6},
  pages={7709--7742},
  year={2023},
  publisher={Springer}
}

@article{EvaluationTextGeneration2020CelikyilmazEtAl,
  title={Evaluation of text generation: A survey},
  author={Celikyilmaz, Asli and Clark, Elizabeth and Gao, Jianfeng},
  journal={arXiv preprint arXiv:2006.14799},
  year={2020}
}

@article{DialogGenerationSurvey2019SanthanamEtAl,
  title={A survey of natural language generation techniques with a focus on dialogue systems-past, present and future directions},
  author={Santhanam, Sashank and Shaikh, Samira},
  journal={arXiv preprint arXiv:1906.00500},
  year={2019}
}

@article{NLGDecoding2021ZarriessEtAl,
  title={Decoding methods in neural language generation: a survey},
  author={Zarrie{\ss}, Sina and Voigt, Henrik and Sch{\"u}z, Simeon},
  journal={Information},
  volume={12},
  number={9},
  pages={355},
  year={2021},
  publisher={MDPI}
}

@article{ThePromptReprt2024SchulhoffEtAl,
  title={The prompt report: a systematic survey of prompt engineering techniques},
  author={Schulhoff, Sander and Ilie, Michael and Balepur, Nishant and Kahadze, Konstantine and Liu, Amanda and Si, Chenglei and Li, Yinheng and Gupta, Aayush and Han, HyoJung and Schulhoff, Sevien and others},
  journal={arXiv preprint arXiv:2406.06608},
  year={2024}
}

@article{StyleTransfer2025YangEtAl,
  title={Steering Large Language Models with Register Analysis for Arbitrary Style Transfer},
  author={Yang, Xinchen and Carpuat, Marine},
  journal={arXiv preprint arXiv:2505.00679},
  year={2025}
}

@inproceedings{Decoding2023Naseh2023,
  title={Stealing the decoding algorithms of language models},
  author={Naseh, Ali and Krishna, Kalpesh and Iyyer, Mohit and Houmansadr, Amir},
  booktitle={Proceedings of the 2023 ACM SIGSAC Conference on Computer and Communications Security},
  pages={1835--1849},
  year={2023}
}

@article{SalientFeatures2024XxuEtAl,
  title={Salient information prompting to steer content in prompt-based abstractive summarization},
  author={Xu, Lei and Karim, Mohammed Asad and Dingliwal, Saket and Elangovan, Aparna},
  journal={arXiv preprint arXiv:2410.02741},
  year={2024}
}

@article{ContextEngineering2025MeiEtAl,
  author       = {Lingrui Mei and
                  Jiayu Yao and
                  Yuyao Ge and
                  Yiwei Wang and
                  Baolong Bi and
                  Yujun Cai and
                  Jiazhi Liu and
                  Mingyu Li and
                  Zhong{-}Zhi Li and
                  Duzhen Zhang and
                  Chenlin Zhou and
                  Jiayi Mao and
                  Tianze Xia and
                  Jiafeng Guo and
                  Shenghua Liu},
  title        = {A Survey of Context Engineering for Large Language Models},
  journal      = {CoRR},
  volume       = {abs/2507.13334},
  year         = {2025},
  url          = {https://doi.org/10.48550/arXiv.2507.13334},
  doi          = {10.48550/ARXIV.2507.13334},
  eprinttype    = {arXiv},
  eprint       = {2507.13334},
  bibsource    = {dblp computer science bibliography, https://dblp.org}
}

@article{ProgramOfThoughts2023ChenEtAl,
  author       = {Wenhu Chen and
                  Xueguang Ma and
                  Xinyi Wang and
                  William W. Cohen},
  title        = {Program of Thoughts Prompting: Disentangling Computation from Reasoning
                  for Numerical Reasoning Tasks},
  journal      = {Trans. Mach. Learn. Res.},
  volume       = {2023},
  year         = {2023},
  url          = {https://openreview.net/forum?id=YfZ4ZPt8zd},
  bibsource    = {dblp computer science bibliography, https://dblp.org}
}

@inproceedings{SelfConsistency2023WangEtAl,
  author       = {Xuezhi Wang and
                  Jason Wei and
                  Dale Schuurmans and
                  Quoc V. Le and
                  Ed H. Chi and
                  Sharan Narang and
                  Aakanksha Chowdhery and
                  Denny Zhou},
  title        = {Self-Consistency Improves Chain of Thought Reasoning in Language Models},
  booktitle    = {The Eleventh International Conference on Learning Representations,
                  {ICLR} 2023, Kigali, Rwanda, May 1-5, 2023},
  publisher    = {OpenReview.net},
  year         = {2023},
  url          = {https://openreview.net/forum?id=1PL1NIMMrw},
  bibsource    = {dblp computer science bibliography, https://dblp.org}
}

@inproceedings{PrefixTuning2021Li,
  author       = {Xiang Lisa Li and
                  Percy Liang},
  editor       = {Chengqing Zong and
                  Fei Xia and
                  Wenjie Li and
                  Roberto Navigli},
  title        = {Prefix-Tuning: Optimizing Continuous Prompts for Generation},
  booktitle    = {Proceedings of the 59th Annual Meeting of the Association for Computational
                  Linguistics and the 11th International Joint Conference on Natural
                  Language Processing, {ACL/IJCNLP} 2021, (Volume 1: Long Papers), Virtual
                  Event, August 1-6, 2021},
  pages        = {4582--4597},
  publisher    = {Association for Computational Linguistics},
  year         = {2021},
  url          = {https://doi.org/10.18653/v1/2021.acl-long.353},
  doi          = {10.18653/V1/2021.ACL-LONG.353},
  bibsource    = {dblp computer science bibliography, https://dblp.org}
}

@inproceedings{TaskArithmetic2023IlharcoEtAl,
  author       = {Gabriel Ilharco and
                  Marco T{\'{u}}lio Ribeiro and
                  Mitchell Wortsman and
                  Ludwig Schmidt and
                  Hannaneh Hajishirzi and
                  Ali Farhadi},
  title        = {Editing models with task arithmetic},
  booktitle    = {The Eleventh International Conference on Learning Representations,
                  {ICLR} 2023, Kigali, Rwanda, May 1-5, 2023},
  publisher    = {OpenReview.net},
  year         = {2023},
  url          = {https://openreview.net/forum?id=6t0Kwf8-jrj},
  bibsource    = {dblp computer science bibliography, https://dblp.org}
}

@article{TaskVectors2024BelanecEtAl,
  author       = {R{\'{o}}bert Belanec and
                  Simon Ostermann and
                  Ivan Srba and
                  M{\'{a}}ria Bielikov{\'{a}}},
  title        = {Task Prompt Vectors: Effective Initialization through Multi-Task Soft-Prompt
                  Transfer},
  journal      = {CoRR},
  volume       = {abs/2408.01119},
  year         = {2024},
  url          = {https://doi.org/10.48550/arXiv.2408.01119},
  doi          = {10.48550/ARXIV.2408.01119},
  eprinttype    = {arXiv},
  eprint       = {2408.01119},
  bibsource    = {dblp computer science bibliography, https://dblp.org}
}

@inproceedings{AutoPrompt2021,
  author       = {Taylor Shin and
                  Yasaman Razeghi and
                  Robert L. Logan IV and
                  Eric Wallace and
                  Sameer Singh},
  title        = {AutoPrompt: Eliciting Knowledge from Language Models with Automatically
                  Generated Prompts},
  booktitle    = {Proceedings of the 2020 Conference on Empirical Methods in Natural
                  Language Processing, {EMNLP} 2020, Online, November 16-20, 2020},
  pages        = {4222--4235},
  publisher    = {Association for Computational Linguistics},
  year         = {2020},
  doi          = {10.18653/V1/2020.EMNLP-MAIN.346}
}

@inproceedings{BlackBoxPromptOpt2023,
  author       = {Jiale Cheng and
                  Xiao Liu and
                  Kehan Zheng and
                  Pei Ke and
                  Hongning Wang and
                  Yuxiao Dong and
                  Jie Tang and
                  Minlie Huang},
  title        = {Black-Box Prompt Optimization: Aligning Large Language Models without
                  Model Training},
  booktitle    = {Proceedings of the 62nd Annual Meeting of the Association for Computational
                  Linguistics (Volume 1: Long Papers), {ACL} 2024, Bangkok, Thailand,
                  August 11-16, 2024},
  pages        = {3201--3219},
  publisher    = {Association for Computational Linguistics},
  year         = {2024},
  doi          = {10.18653/V1/2024.ACL-LONG.176},
}

@inproceedings{ShiContrastiveDecoding2023,
  title={Trusting your evidence: Hallucinate less with context-aware decoding},
  author={Shi, Weijia and Han, Xiaochuang and Lewis, Mike and Tsvetkov, Yulia and Zettlemoyer, Luke and Yih, Wen-tau},
  booktitle={Proceedings of the 2024 Conference of the North American Chapter of the Association for Computational Linguistics: Human Language Technologies (Volume 2: Short Papers)},
  pages={783--791},
  year={2024}
}

@inproceedings{Zheng2023LLMJudge,
  author       = {Lianmin Zheng and
                  Wei{-}Lin Chiang and
                  Ying Sheng and
                  Siyuan Zhuang and
                  Zhanghao Wu and
                  Yonghao Zhuang and
                  Zi Lin and
                  Zhuohan Li and
                  Dacheng Li and
                  Eric P. Xing and
                  Hao Zhang and
                  Joseph E. Gonzalez and
                  Ion Stoica},
  title        = {Judging LLM-as-a-Judge with MT-Bench and Chatbot Arena},
  booktitle    = {Advances in Neural Information Processing Systems 36: Annual Conference
                  on Neural Information Processing Systems 2023, NeurIPS 2023, New Orleans,
                  LA, USA, December 10 - 16, 2023},
  year         = {2023},
}

@article{SurverLLMAsJudge2024GuEtAl,
  author       = {Jiawei Gu and
                  Xuhui Jiang and
                  Zhichao Shi and
                  Hexiang Tan and
                  Xuehao Zhai and
                  Chengjin Xu and
                  Wei Li and
                  Yinghan Shen and
                  Shengjie Ma and
                  Honghao Liu and
                  Yuanzhuo Wang and
                  Jian Guo},
  title        = {A Survey on LLM-as-a-Judge},
  journal      = {CoRR},
  volume       = {abs/2411.15594},
  year         = {2024},
  url          = {https://doi.org/10.48550/arXiv.2411.15594},
  doi          = {10.48550/ARXIV.2411.15594},
  eprinttype    = {arXiv},
  eprint       = {2411.15594},
  bibsource    = {dblp computer science bibliography, https://dblp.org}
}

@article{Lu2021NeuroLogic,
  title={Neurologic decoding:(un) supervised neural text generation with predicate logic constraints},
  author={Lu, Ximing and West, Peter and Zellers, Rowan and Bras, Ronan Le and Bhagavatula, Chandra and Choi, Yejin},
  journal={arXiv preprint arXiv:2010.12884},
  year={2020}
}

@article{Su2022Contrastive,
  title={Contrastive search is what you need for neural text generation},
  author={Su, Yixuan and Collier, Nigel},
  journal={arXiv preprint arXiv:2210.14140},
  year={2022}
}

@article{long2023ToT,
  title={Large language model guided tree-of-thought},
  author={Long, Jieyi},
  journal={arXiv preprint arXiv:2305.08291},
  year={2023}
}

@inproceedings{EvaluLLM2024PanEtAl,
    title = "Human-Centered Design Recommendations for {LLM}-as-a-judge",
    author = "Pan, Qian  and
      Ashktorab, Zahra  and
      Desmond, Michael  and
      Santill{\'a}n Cooper, Mart{\'i}n  and
      Johnson, James  and
      Nair, Rahul  and
      Daly, Elizabeth  and
      Geyer, Werner",
    booktitle = "Proceedings of the 1st Human-Centered Large Language Modeling Workshop",
    year = "2024",
    publisher = "ACL",

    doi = "10.18653/v1/2024.hucllm-1.2",
    pages = "16--29",
   
}

@article{SclarEtAl2023Quantifying,
  title={Quantifying Language Models' Sensitivity to Spurious Features in Prompt Design or: How I learned to start worrying about prompt formatting},
  author={Sclar, Melanie and Choi, Yejin and Tsvetkov, Yulia and Suhr, Alane},
  journal={arXiv preprint arXiv:2310.11324},
  year={2023}
}

@article{LiEtAl2024LlmsAsAJudge,
  title={Llms-as-judges: a comprehensive survey on llm-based evaluation methods},
  author={Li, Haitao and Dong, Qian and Chen, Junjie and Su, Huixue and Zhou, Yujia and Ai, Qingyao and Ye, Ziyi and Liu, Yiqun},
  journal={arXiv preprint arXiv:2412.05579},
  year={2024}
}

@article{NovikovaEtAl2025Consistency,
  title={Consistency in language models: Current landscape, challenges, and future directions},
  author={Novikova, Jekaterina and Anderson, Carol and Blili-Hamelin, Borhane and Rosati, Domenic and Majumdar, Subhabrata},
  journal={arXiv preprint arXiv:2505.00268},
  year={2025}
}

@article{PatelEtAl2025Optimization,
  title={Towards Interpretable Soft Prompts},
  author={Patel, Oam and Wang, Jason and Nayak, Nikhil Shivakumar and Srinivas, Suraj and Lakkaraju, Himabindu},
  journal={arXiv preprint arXiv:2504.02144},
  year={2025}
}

@article{VatsalEtAl2025Multilingual,
  title={Multilingual Prompt Engineering in Large Language Models: A Survey Across NLP Tasks},
  author={Vatsal, Shubham and Dubey, Harsh and Singh, Aditi},
  journal={arXiv preprint arXiv:2505.11665},
  year={2025}
}

\end{document}